\documentclass{article}

\usepackage{microtype}
\usepackage{graphicx}
\usepackage{subfigure}
\usepackage{booktabs} 

\usepackage{hyperref}

\usepackage[accepted]{icml2023}

\usepackage{amsmath}
\usepackage{amssymb}
\usepackage{mathtools}
\usepackage{amsthm}

\usepackage[capitalize,noabbrev]{cleveref}

\theoremstyle{plain}

\theoremstyle{definition}

\theoremstyle{remark}

\usepackage[textsize=tiny]{todonotes}

\usepackage{bm}

\icmltitlerunning{Universal Morphology Control via Contextual Modulation}

\begin{document}

\twocolumn[
\icmltitle{Universal Morphology Control via Contextual Modulation}



\icmlsetsymbol{equal}{*}

\begin{icmlauthorlist}
\icmlauthor{Zheng Xiong}{ox}
\icmlauthor{Jacob Beck}{ox}
\icmlauthor{Shimon Whiteson}{ox}
\end{icmlauthorlist}

\icmlaffiliation{ox}{Department of Computer Science, University of Oxford, Oxford, United Kingdom}

\icmlcorrespondingauthor{Zheng Xiong}{zheng.xiong@cs.ox.ac.uk}

\icmlkeywords{Machine Learning, ICML}

\vskip 0.3in
]



\printAffiliationsAndNotice{} 

\begin{abstract}
Learning a universal policy across different robot morphologies can significantly improve learning efficiency and generalization in continuous control. 
However, it poses a challenging multi-task reinforcement learning problem, as the optimal policy may be quite different across robots and critically depend on the morphology. 
Existing methods utilize graph neural networks or transformers to handle heterogeneous state and action spaces across different morphologies, but pay little attention to the dependency of a robot's control policy on its morphology context. 
In this paper, we propose a hierarchical architecture to better model this dependency via contextual modulation, which includes two key submodules: 
(1) Instead of enforcing hard parameter sharing across robots, we use hypernetworks to generate morphology-dependent control parameters; 
(2) We propose a fixed attention mechanism that solely depends on the morphology to modulate the interactions between different limbs in a robot. 
Experimental results show that our method not only improves learning performance on a diverse set of training robots, but also generalizes better to unseen morphologies in a zero-shot fashion. 
The code is publicly available at \url{https://github.com/MasterXiong/ModuMorph}. 
\end{abstract}

\section{Introduction}

Reinforcement learning (RL) for robotic control has made great progress in recent years \citep{levine2016end,kalashnikov2018scalable,andrychowicz2020learning,brohan2022rt}. 
However, the control policy learned on one robot usually cannot transfer to another robot with a different morphology due to their incompatible state and action spaces. 
Given the huge number of possible robot morphologies and the high sample complexity of RL, the currently dominant paradigm of learning a new policy from scratch for each robot morphology is not scalable, and universal controllers that can generalize across different morphologies are desirable to improve learning efficiency, i.e., we want to learn a universal controller with much less environment interactions compared to the total samples required to learn a separate controller for each robot to control. 

Multi-task RL (MTRL) \citep{vithayathil2020survey} provides a promising solution to this challenge by treating the control of each robot as a unique task. 
Instead of learning a separate policy for each morphology, MTRL learns a single policy, conditioned on both the robot state and the morphology, to generalize across different robots. 
From this MTRL perspective, the robot morphology is important context that helps identify the task, as the optimal control policy of a robot critically depends on its morphology. 
For example, if an animal injures a leg, which changes its morphology, then a different gait may be required for locomotion.
Similarly, an animal's tail
can significantly influence its locomotion even if all the other body parts remain unchanged \citep{jagnandan2017lateral}. 

However, previous work on universal morphology control mainly focuses on policy architecture design, such as using graph neural networks (GNNs) \citep{wang2018nervenet,huang2020one} or transformers \citep{kurin2021my,gupta2022metamorph}, to enable generalization over heterogeneous state and action spaces, as the number of limbs differs across morphologies. 
By contrast, little attention has been paid to how to effectively utilize the morphology context in the control policy. 
While some recent works propose to feed morphology context as an additional input to the policy network \citep{gupta2022metamorph}, or add a morphology-aware positional encoding (PE) to the state representation \citep{gupta2022metamorph,hong2022structureaware}, in effect they are equivalent to just adding a context-conditioned bias term to the node embedding layer in the network. 
This may lack sufficient model capacity to represent the diverse policies required to control different morphologies, as supported by both theoretical \citep{galanti2020modularity} and empirical evidence \citep{ben2022context,beck2022hypernetworks} from previous work in multi-task learning and meta-learning. 

To better utilize task context for morphology control, we propose a hierarchical policy architecture consisting of a base controller, and a context modulator that regulates the control policy according to the characteristics of different morphologies. 
We name our method as \emph{ModuMorph} to highlight its architecture novelty in contextual modulation. 
Specifically, ModuMorph includes two submodules. 
First, we modulate network parameters in the base controller with hypernetworks (HN) \citep{ha2016hypernetworks}. 
Conditioned on the morphology context, HN can generate different policy parameters for different robots, which helps improve behavior diversity across morphologies. 
Second, we modulate the attention weight matrices in the transformer layers of the base controller with morphology context alone, which introduces a structure-aware inductive bias on how each limb in a robot should attend to the others to update its own behaviors. 

In principle, the proposed contextual modulator can be incorporated into any transformer-based architectures for morphology control, while the HN module can also work with GNN-based architectures. 
In this paper, we use a recently proposed transformer-based method, MetaMorph \citep{gupta2022metamorph}, as the backbone algorithm for experiments due to its superior performance and efficient implementation. 
Our experiments on a challenging morphology control benchmark called UNIMAL \citep{gupta2021embodied}, which includes hundreds of diverse morphologies, show that using contextual modulation improves not only the learning performance on training morphologies, but also the zero-shot generalization performance on unseen test morphologies, which validates the effectiveness of our method. 

\section{Background}
\subsection{Problem Formulation}
Consider the problem of learning a universal policy to control a set of $K$ robots with different morphologies. 
For each robot $k$, the control problem can be seen as a contextual Markov Decision Process (CMDP) \citep{hallak2015contextual} defined as a tuple $(\mathcal{S}_k, \mathcal{A}_k, \mathcal{C}_k, T_k, R_k)$, where $\mathcal{S}_k$, $\mathcal{A}_k$, $\mathcal{C}_k$, $T_k$, $R_k$ are the state space, action space, task context, transition function and reward function respectively. 

We assume that all the robots are drawn from a modular design space, i.e., each robot can be seen as a morphology tree over a set of basic nodes (limbs), and all the nodes share the same node-level state and action space. 
Based on this assumption, we have $ \mathcal{S}_k = \{ \mathcal{S}_k^i | i=1, \dots, N_k \}$ and $ \mathcal{A}_k = \{ \mathcal{A}_k^i | i=1, \dots, N_k \}$, where $N_k$ is the number of nodes in robot $k$. 
The task context includes morphology information about the robot, consisting of node-wise context $\{ \mathcal{C}_k^i | i=1, \dots, N_k \}$ (such as the size and mass of the limb and its initial position relative to its parent node), and an adjacency matrix that defines the topology of the morphology tree. 

We use $s_{k,t}, a_{k,t}, r_{k,t}$ to represent the state, action and reward at time step $t$ for robot $k$. 
The training objective is to learn a universal policy $\pi_\theta(a_{k,t}|s_{k,t}, c_k)$ to maximize the average return over all the training morphologies, i.e., $ \max_{\theta} \left[ \frac{1}{K} \sum_{k=1}^K \sum_{t=0}^H r_{k,t} \right] $, where $H$ is the task horizon for all different robots. 
In addition to good training performance, we also expect the learned policy to generalize well on unseen test morphologies in a zero-shot manner. 

\subsection{Transformers for Universal Morphology Control}
Transformers \citep{vaswani2017attention} can model the interactions between a set of elements of arbitrary size, thus are well suited to process different morphologies with various number of limbs. 

For morphology control, the attention module in transformers determines how each node attends to the others to update its own node representation. 
It requires three input vectors from each node $i$, i.e., a query $\bm{q}_i$, a key $\bm{k}_i$ of dimension $d_k$, and a value vector $\bm{v}_i$ of dimension $d_v$. 
The three vectors for each node are stacked into matrices $Q, K, V$, and the attention module updates the node representation as $$ \text{Attention}(Q,K,V) = \text{softmax}(\frac{QK^T}{\sqrt{d_k}})V, $$
i.e., the dot product of $\bm{q}_i$ and $\bm{k}_j$ determines how much attention node $i$ pays to node $j$ to update its node representation. 
Usually multiple attention heads are trained independently to learn different node interactions. 
The attention block is then followed by feedforward layers to form a whole transformer module, with normalization and skip connection operations in between. 
Several layers of transformer modules can be stacked to further improve model capacity. 

\subsubsection{MetaMorph}
\begin{figure}
    \centering
    \includegraphics[width=\columnwidth]{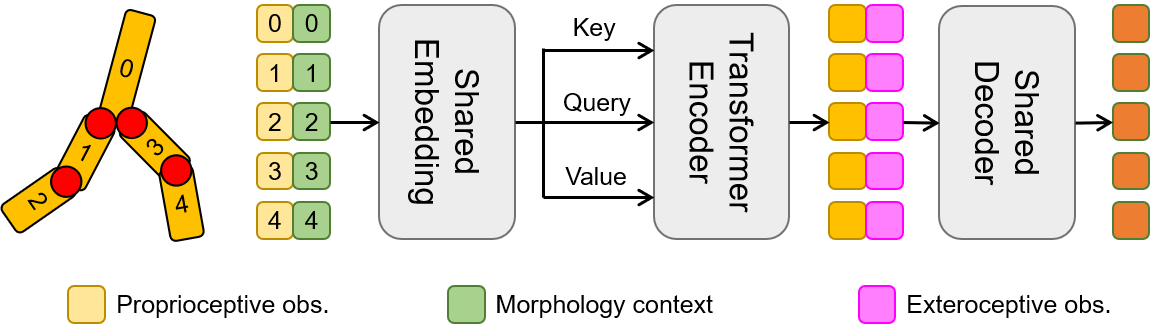}
    \caption{The framework of MetaMorph \citep{gupta2022metamorph}. On this two-leg robot for example, its nodes are ordered by depth-first tree search, with the torso node as the tree root. MetaMorph concatenates proprioceptive observations and morphology context as node inputs, processes them with a shared embedding layer, a transformer encoder and a shared decoder sequentially. Exteroceptive observations are concatenated as decoder inputs if needed. }
    \label{fig:metamorph}
\end{figure}

MetaMorph \citep{gupta2022metamorph} is a transformer-based method for universal morphology control (Figure \ref{fig:metamorph}). 
It concatenates the time-variant proprioceptive observation $s_{k,t}^i$ and the time-invariant morphology context $c_k^i$ as the input vector of node $i$. 
The node input first goes through an embedding layer shared across all nodes to get node embedding $\bm{e}_i$. 
Then the node embeddings are updated by a transformer encoder. 
The key, query and value inputs to the transformer are all determined by the node embedding, i.e., $ \bm{k}_i=W_k \bm{e}_i, \bm{q}_i=W_q \bm{e}_i, \bm{v}_i=W_v \bm{e}_i $, where $W_k, W_q, W_v$ are learnable weight matrices. 
After transformer encoding, if there are globally exteroceptive observations, such as a height map of the agent's surroundings in a changing terrain, then they are processed by a multi-layer perceptron (MLP) and concatenated to the node features. 
Incorporating exteroceptive observations is essential to enable the agent to take different actions in different environmental conditions. 
Finally, the concatenated features go through a decoder shared across all nodes to generate the actions for each node. 

In MetaMorph, the morphology context $c_k$ is only utilized as an additional node input, which is equivalent to adding a context-conditioned bias to the node embedding, as $c_k$ remains unchanged on each robot. 
However, the optimal control policy can significantly vary across robots. 
Simply adding context-conditioned bias terms to the node embeddings, while sharing all the other model parameters, thus may not have sufficient expressive power to represent the diverse policies required for different morphologies \citep{galanti2020modularity,ben2022context,beck2022hypernetworks}. 
To tackle this limitation, we propose two contextual modulation approaches to learn more diverse context-conditioned policies across different morphologies in Section \ref{sec:method}. 

MetaMorph also adds a learned positional encoding (PE) to the node embedding, i.e., $\bm{e}_i = \text{Encoder}(s_{k,t}^i, c_k^i) + \text{PE}_i$, where $\text{PE}_i$ is a learnable vector that is shared across all the nodes with index $i$ across different morphologies. 
PE is a common way to inject positional information back into transformers, as the attention module alone is order-invariant \citep{vaswani2017attention,dufter2022position}. 
However, we find that PE actually provides little help in universal morphology control and thus omit it in Figure \ref{fig:metamorph} for simplicity. We analyze why it does not work in Appendix \ref{appenfix:metamorph}. 

\subsection{Hypernetworks}

A hypernetwork \citep{ha2016hypernetworks} is a network that generates the parameters of a base network $\theta$ conditioned on some meta variables $c$, i.e., $\theta=\text{HN}_\phi(c)$, where $\phi$ is the HN parameters to learn. 
Under the MTRL setting, the meta variables correspond to the task context. 
With HN, $\theta$ turns into a context-dependent function that may better reflect the dependency between the task context and the base network's parameters. 
Compared to the common practice of integrating task context into the base network by concatenating it to the base network's input vector, modeling their relationships via HN enjoys better parameter complexity \citep{galanti2020modularity} and lower gradient variance during learning \citep{sarafian2021recomposing}. 
However, HN is also known to be harder to optimize due to its more complicated hierarchical network architecture, and proper initialization of HN is critical to stabilize its training \citep{chang2019principled,beck2022hypernetworks}. 

\section{Universal Morphology Control via Contextual Modulation}
\label{sec:method}

\begin{figure}
    \centering
    \includegraphics[width=\columnwidth]{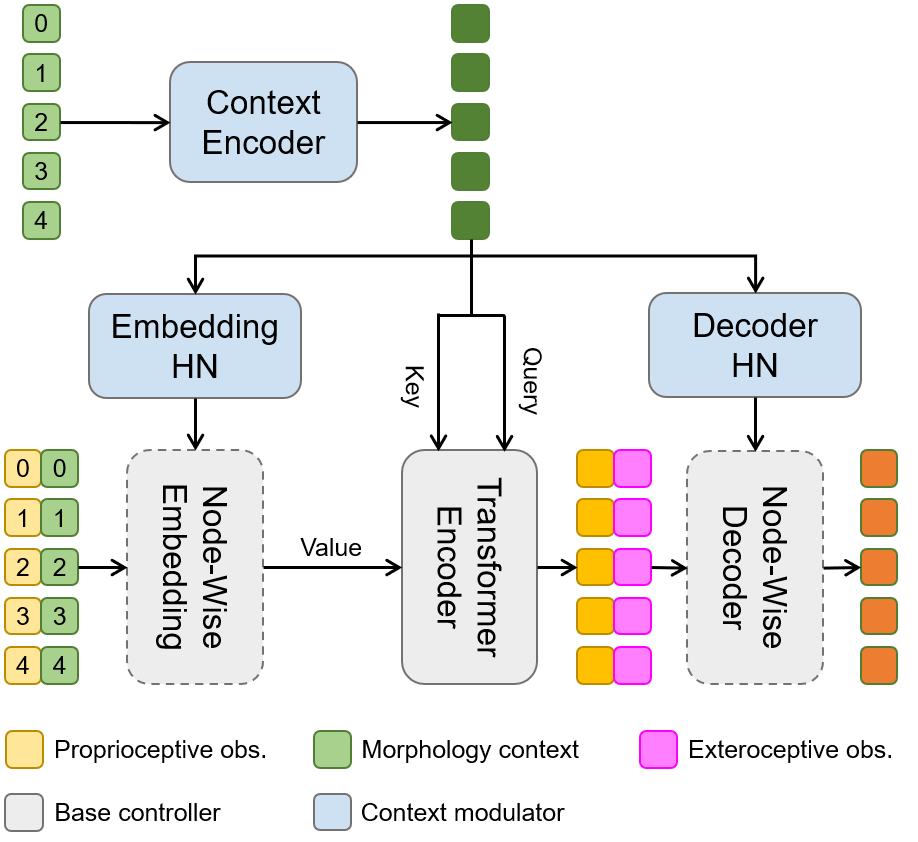}
    \caption{The hierarchical framework of our proposed method. The morphology context modulates the base controller in two ways: (1) Generating context-conditioned embedding and decoder parameters via an HN. We use dotted edges to highlight that these two modules are not shared across different nodes and morphologies as in MetaMorph. (2) Generating morphology-conditioned attention matrices by using context embeddings as the key and query inputs to the transformer encoder. Note that we use two separate context encoders for these two submodules, but show only a single shared context encoder in this figure for ease of illustration. }
    \label{fig:framework}
\end{figure}

In this section, we introduce two novel approaches to modulate the controller with morphology context. 
The framework of our proposed method is shown in Figure \ref{fig:framework}, which includes a base controller that generally follows the same architecture as MetaMorph, and a context network that modulates the base controller in two ways: 
(1) Instead of using shared embedding layer and decoder across all nodes, we generate node-wise embedding and decoder parameters with an HN conditioned on the morphology context. 
(2) The node embedding in the base controller is only used to generate the value input to the transformer encoder, while the key and query are conditioned on the morphology context to generate a fixed attention matrix. 
We call these two approaches hypernetworks (HN) and fixed attention (FA).

\subsection{Context Conditioning via Hypernetworks}
\label{subsec:HN}

While GNN and transformer-based controllers enable generalization across different morphologies, they also introduce a structural constraint that all nodes, both within a single robot and across different morphologies, have to share the same modular control policy. 
This hard parameter sharing mechanism \citep{ruder2017overview} may limit the behavior diversity across different nodes and the controller's model capacity to learn the optimal policy, as we usually expect different limbs to follow different control strategies based on their roles in the morphology. 
There is even evidence from neuroscience that the muscles in human body are controlled by different types of motor neurons based on their identities \citep{stifani2014motor}. 

Consequently, we hypothesize that instead of learning parameters shared across all nodes, learning node-wise parameters may improve behavior diversity and learning performance. 
We first conduct a proof-of-concept experiment to validate our hypothesis, then show how to generate context-conditioned parameters for each node via HN modulation. 

\subsubsection{A Proof-of-Concept Experiment}
\begin{figure}[t]
    \centering
    \includegraphics[width=\columnwidth]{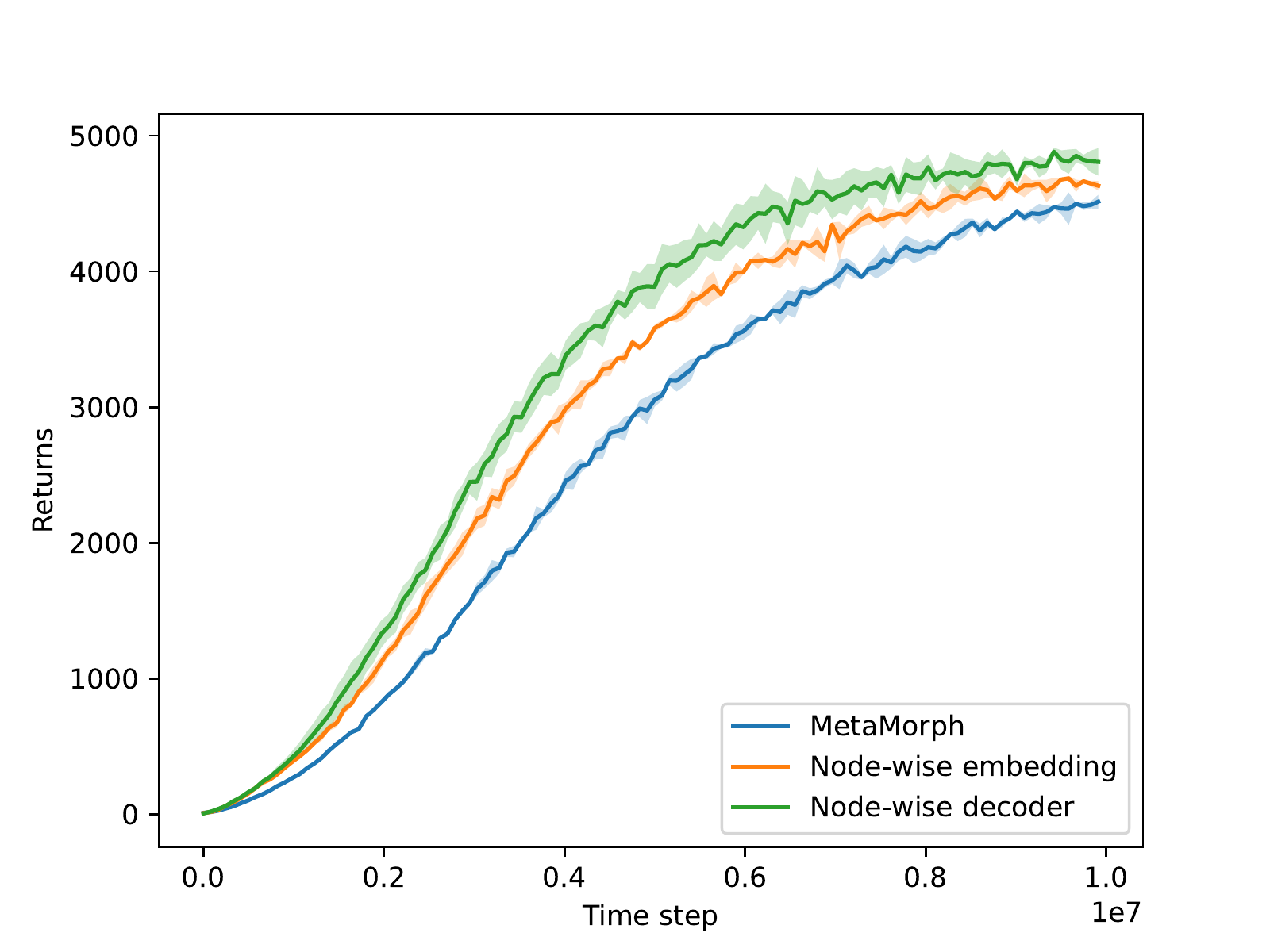}
    \caption{Single-robot learning curves averaged over 20 robots. }
    \label{fig:motivation_ST}
\end{figure}
We design a motivating experiment to show that enabling behavior diversity across nodes via learning node-wise parameters 
can improve training performance. 

We train a MetaMorph model on a single robot. 
However, instead of learning a single embedding layer shared across all the nodes, we learn a separate embedding for each node. 
Similarly, we train another MetaMorph variant with node-wise decoders. 
We randomly sample 20 morphologies from the UNIMAL benchmark and run single-task training on each of them for 10M steps. 
As shown in Figure \ref{fig:motivation_ST}, the node-wise embedding and node-wise decoder variants both outperform MetaMorph, which shows that enabling behavior diversity across the nodes of a single robot is helpful. 

However, the approach used in our proof-of-concept experiment is impractical for two reasons: (1) It cannot generalize to new morphologies unseen during training; (2) Learning separate parameters for each node is not scalable, as the number of parameters to learn grows linearly with the number of morphologies. 
Consequently, we next introduce HN modulation to enable behavior diversity while maintaining generalization and scalability of the learned model. 

\subsubsection{Context-Conditioned Parameter Generation via Hypernetworks}

Learning separate parameters for each node has generalization and scalability issues, as it does not utilize the similarity between nodes. 
Intuitively, if two nodes play similar roles in two different morphologies (e.g., they are both the left thigh in their robots), then we may expect them to also have similar node-wise parameters. 
As the morphology context of each node can provide rich information about the similarities between nodes, we propose to generate node-wise parameters via a context-conditioned HN, i.e., $\theta_k^i = \text{HN}_\phi(c_k^i) $, where $\theta_k^i$ is the node-wise parameters for node $i$ of robot $k$, and $\text{HN}_\phi$ is the learned HN shared across all nodes. 
Generating node-wise parameters via HN is scalable, as we only need to additionally learn one set of HN parameters $\phi$ regardless of the number of morphologies, and generalizable, as we can directly feed new node context on unseen morphologies into the HN to generate its control parameters. 

To better illustrate how HN-generated parameters work, we take the node embedding layer as an example. 
In MetaMorph, the embedding layer consists of a single set of weights $W$ and bias $b$ shared across all the nodes, i.e., $e_k^i = W x_k^i + b$, where $x_k^i$ and $e_k^i$ are the node input and node embedding of node $i$ in robot $k$ respectively. 
For our HN approach, however, the embedding layer's parameters are different across nodes, i.e., $e_k^i = W_k^i x_k^i + b_k^i$, where $W_k^i = \text{HN}_W(c_k^i)$ and $b_k^i = \text{HN}_b(c_k^i)$ are node-wise weights and bias generated by HN conditioned on the node context. 

In practice, we only generate linear layers' parameters in the base network with HN, i.e., the embedding layer and the decoder. 
The transformer encoder is still shared across all morphologies, as there are too many weight matrices in a transformer layer to efficiently generate them via HN. 

\subsection{Morphology-Conditioned Fixed Attention}
\label{subsec:FA}

The attention weight matrix plays an important role in transformers as it determines how each node should attend to the others to update its own representation. 

Existing methods use the node embedding in the base controller as the key and query inputs to generate the attention weights, which change dynamically at every time step due to the time-variant proprioceptive observations. 
However, determining attention weights in such a dynamic way may not well reflect how different nodes interact. 
Instead, it may be the case that the attention of one node to the others should depend \emph{solely} on the morphology of the agent. 
For example, when you want to grasp an object within your reach, you pay more attention to the state of your arm than your leg to determine the movement of your hand. 
Similarly, whether you are standing or sitting, which changes the proprioceptive observations of body parts, has little influence on your attention strategy for grasping. 

Consequently, we hypothesize that it may be beneficial to incorporate such intuitions as inductive biases into the controller architecture, i.e., each node should attend to the other nodes in a static way, and the attention weights should be determined by the morphology context alone. 
To realize these inductive biases, we pass the node context through a context encoder, and use the context embedding as the key and query to modulate the transformer in the base controller, while the node embedding in the base network is only used as the value input (Figure \ref{fig:framework}). 
As the morphology context remains unchanged, the attention matrix is fixed on each robot to reflect the structural relationships between nodes. 

\subsection{Computational Cost}
HN learning will increase computational cost during training. 
However, there is no additional cost during deployment, as we can generate node-wise parameters with HN on each robot in advance. 
On the other hand, except for context encoding, FA will introduce no additional computation during training, as it just changes the query and key inputs to the transformer. 
Furthermore, FA could even reduce the computation during evaluation, as we need to compute the FA weights only once for each robot and then can reuse it afterwards.  
See Appendix \ref{appendix:implementations} for more implementation details of our contextual modulation method. 

\section{Experimental Setup}
\begin{figure*}[t]
    \centering
    \includegraphics[width=\textwidth]{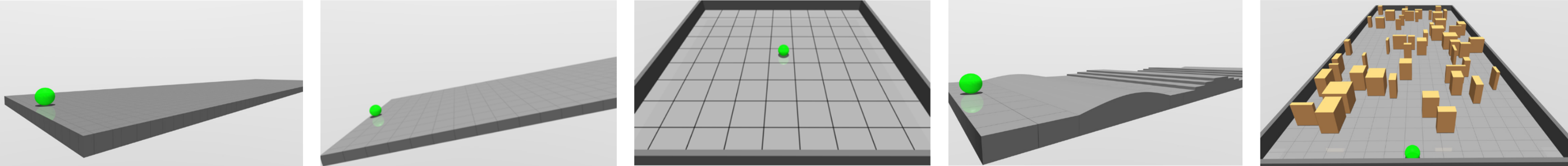}
    \caption{The five environments used for experiments. From left to right: Flat terrain (FT), Incline, Exploration, Variable terrain (VT), Obstacles. Images credit to \citet{gupta2021embodied,gupta2022metamorph}. }
    \label{fig:environments}
\end{figure*}
\paragraph{Environments}
We experiment on the UNIMAL task set as used in MetaMorph \citep{gupta2022metamorph}, which includes 100 training robots and 100 test robots with diverse morphologies \citep{gupta2021embodied}. 

We consider five different environments from \citet{gupta2021embodied} for our experiments (Figure \ref{fig:environments}): 
(1) Flat terrain (FT): maximize locomotion distance on a flat floor; 
(2) Incline: maximize locomotion distance on an incline of 10 degrees; 
(3) Exploration: maximize the number of distinct grids visited on a flat arena discretized into grids; 
(4) Variable terrain (VT): maximize locomotion distance on a variable terrain with three different terrain types. For each episode, a new terrain is generated by randomly sampling a sequence of terrain types and interleaving them with flat terrain. 
(5) Obstacles: maximize locomotion distance on a flat terrain with randomly positioned obstacles. 
The first three environments only require proprioceptive observations and morphology context as model input, while the last two require height map information surrounding the robot as additional exteroceptive observation input to the controller, so that the agent can perceive and react to different terrains or obstacles in its way. 

\paragraph{Baselines}
We consider both multi-robot (MR) and single-robot (SR) baselines. 

For MR training, we use MetaMorph \citep{gupta2022metamorph} as the baseline. 
However, we notice two issues in the MetaMorph code and thus implement a slightly modified version to eliminate these issues. 
We name the modified version as MetaMorph*, and build our modulation modules upon it. 
In general, MetaMorph* achieves similar or even better performance compared to MetaMorph in most environments, and we report the results of both for a fair comparison. 
See Appendix \ref{appenfix:metamorph} for more details on the difference between MetaMorph and MetaMorph*. 

For SR training, we train an MLP policy on each robot, and consider two different training budgets for different purposes. 
First, we do SR training with the same per-robot budget as in MR training, which is named as \emph{SR-fair} and used to compare the sample efficiency of MR and SR learning. 
Second, We do SR training for 10M steps on each robot till convergence, which is named as \emph{SR-10M} and used as a performance upper bound. 
We choose to use an MLP of 3 hidden layers, each with 256 hidden units by performing a grid search over the layer number and hidden size. 

\paragraph{Ablations}
As we propose two different approaches for contextual modulation, we test ablations by adding only HN or FA to the MetaMorph* baseline, and compare them with the full version of adding both to MetaMorph*. 

\paragraph{Training Setup}
We train for 100M steps in FT, Incline, and Exploration, and 200M steps in VT and Obstacles, as they are more challenging to solve due to variable terrains. 
We run three random seeds for each method in each environment, and report the average performance and standard deviation. 
Following the same setup as in MetaMorph, we use PPO \citep{schulman2017proximal} as the optimization algorithm. 
Similar to previous works \citep{dossa2021empirical,sun2022you}, we notice that the early stopping threshold has a significant influence on PPO performance (see Appendix \ref{appenfix:early_stop}).
We thus tune this hyperparameter over the candidate set of $\{ 0.03, 0.05 \}$ for each method in each environment. 
All the remaining hyperparameters follow the same setup as in MetaMorph for a fair comparison. 

\paragraph{Evaluation Setup}
We evaluate zero-shot generalization to unseen robots under two settings with increasing difficulties. 
First, we test on new robots that have the same topology as the training ones but differ in kinematics or dynamics parameters. 
For each parameter to test, we create 4 variants of each training robot by randomly changing the value of the corresponding parameter on all the limbs. 
Second, we evaluate zero-shot generalization to new morphologies which have different topology graphs compared to those seen during training. 
The robots used for both settings are adopted from \citet{gupta2022metamorph} for a fair comparison. 
For each robot, we collect 64 rollouts with randomly sampled initial states. 
We use the average episodic return over the test morphologies to measure the policy's transferability. 

\section{Results}

\subsection{Training Results}
\begin{figure*}[t]
    \centering
    \includegraphics[width=\textwidth]{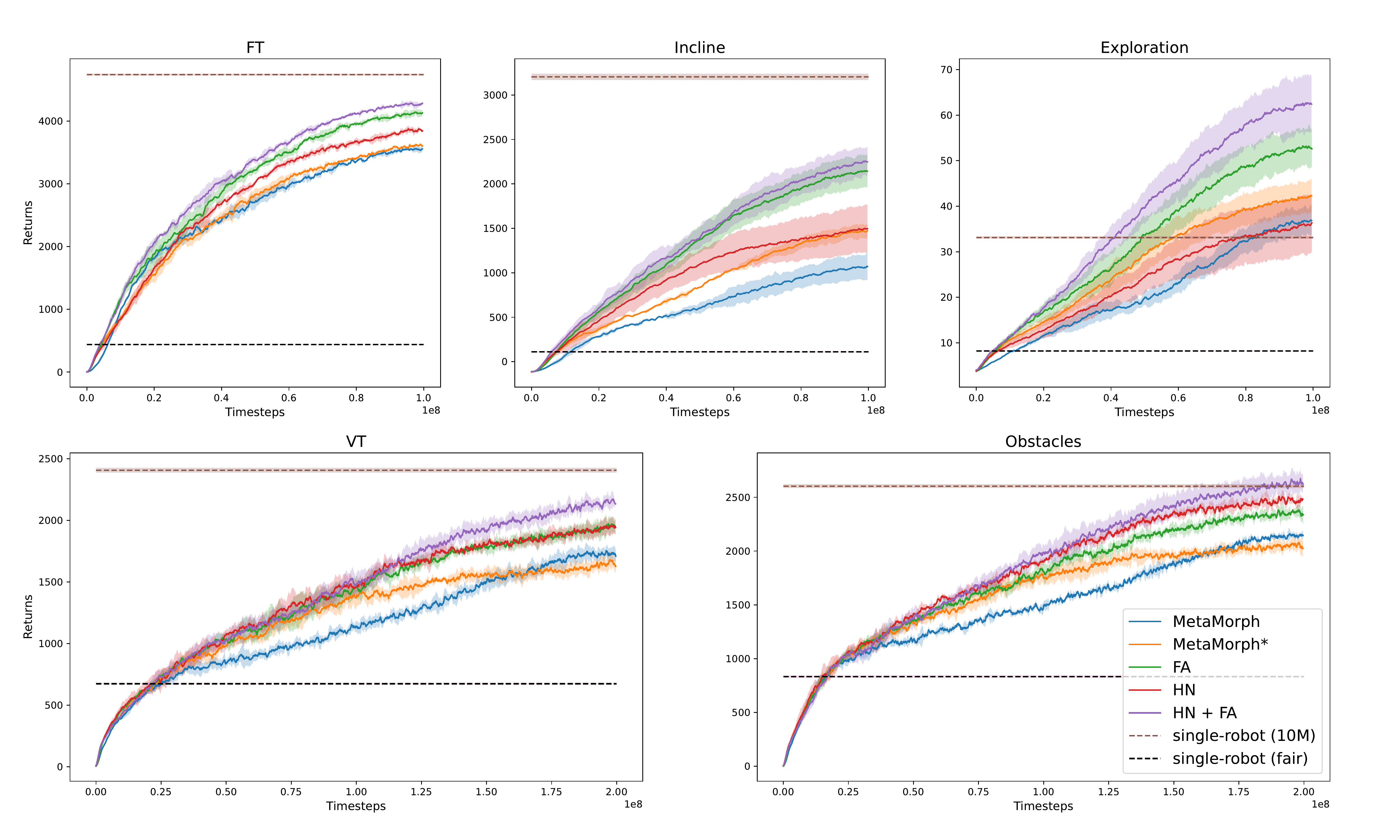}
    \caption{The training curves of different methods in each environment.}
    \label{fig:training_curves}
\end{figure*}
As shown in Figure \ref{fig:training_curves}, all MR methods significantly outperform SR-fair, illustrating the advantage of MTRL in sample efficiency. 
However, there is still a clear performance gap between the two MR baselines and SR-10M, due to the challenges of MTRL. 
Our method significantly reduces this gap (even outperforms SR-10M in Exploration), and consistently outperforms the two MR baselines in all the five environments w.r.t.\ both learning efficiency and final performance. 
Compared to MetaMorph*, which our method builds upon, contextual modulation improves the final performance by $19\%$, $53\%$, $48\%$, $31\%$ and $29\%$ in each environment respectively. 
Although MetaMorph outperforms MetaMorph* in VT and Obstacles, our method still consistently outperforms MetaMorph in these two environments. 

Ablation results show that both FA and HN contribute to the effectiveness of our method. 
FA consistently improves upon MetaMorph* in all the environments, and seems to be more effective in the three environments with unchanged terrain (VT, Incline and Exploration). 
On the other hand, HN helps in three of the five environments, and contributes more in the two environments with changing terrains (VT and Obstacles). 
Next we give some more detailed analysis on the two submodules of our method. 

\paragraph{Fixed Attention}
FA introduces a strong inductive bias that how each node attends to the others should be solely determined by the morphology, and is proved to be effective in all the five environments. 
However, in environments with changing terrains, the robot may need to adopt different gaits in different terrains. 
While in principle this can be realized by taking terrain information as additional decoder input, an alternative idea is to further condition the attention weights on the terrain information, so that the nodes can attend to each other with dynamic terrain-conditioned weights to realize different gaits. 
We thus tried adding the height map as an additional input to compute attention weights, but got results even worse than the MetaMorph* baseline. 
The reason might be that the robot can already adapt to different terrains by taking the height map as decoder input, so the attention module only needs to model intra-morphology interactions, while using terrain info as attention inputs may introduce further optimization challenges. 
Nevertheless, it remains an interesting open question whether learning performance can be further improved by properly incorporating terrain information into attention computation. 

\paragraph{Hypernetworks}
HN provides more significant improvement in the more challenging environments of VT and Obstacles with variable terrains. 
This may imply that behavior diversity across nodes is more important in environments that require complex locomotion skills, while in easier terrains, the benefits of HN may be outweighed by its optimization challenges. 
Moreover, adding HN harms learning performance in the Exploration environment. 
The training statistics show that the HN variant has a much higher error in value prediction compared to the other methods in Exploration, which may be the reason for its worse performance. 
Value prediction is particularly hard in Exploration, as the value depends on not only the robot's status, but also the robot's visitation history in the arena, which is not accessible to the robot. 
We hypothesize that this problem is more severe when using the more complex HN architecture. 

\subsection{Zero-Shot Generalization to Kinematics and Dynamics Variations}
\begin{figure}[t]
    \centering
    \includegraphics[width=\columnwidth]{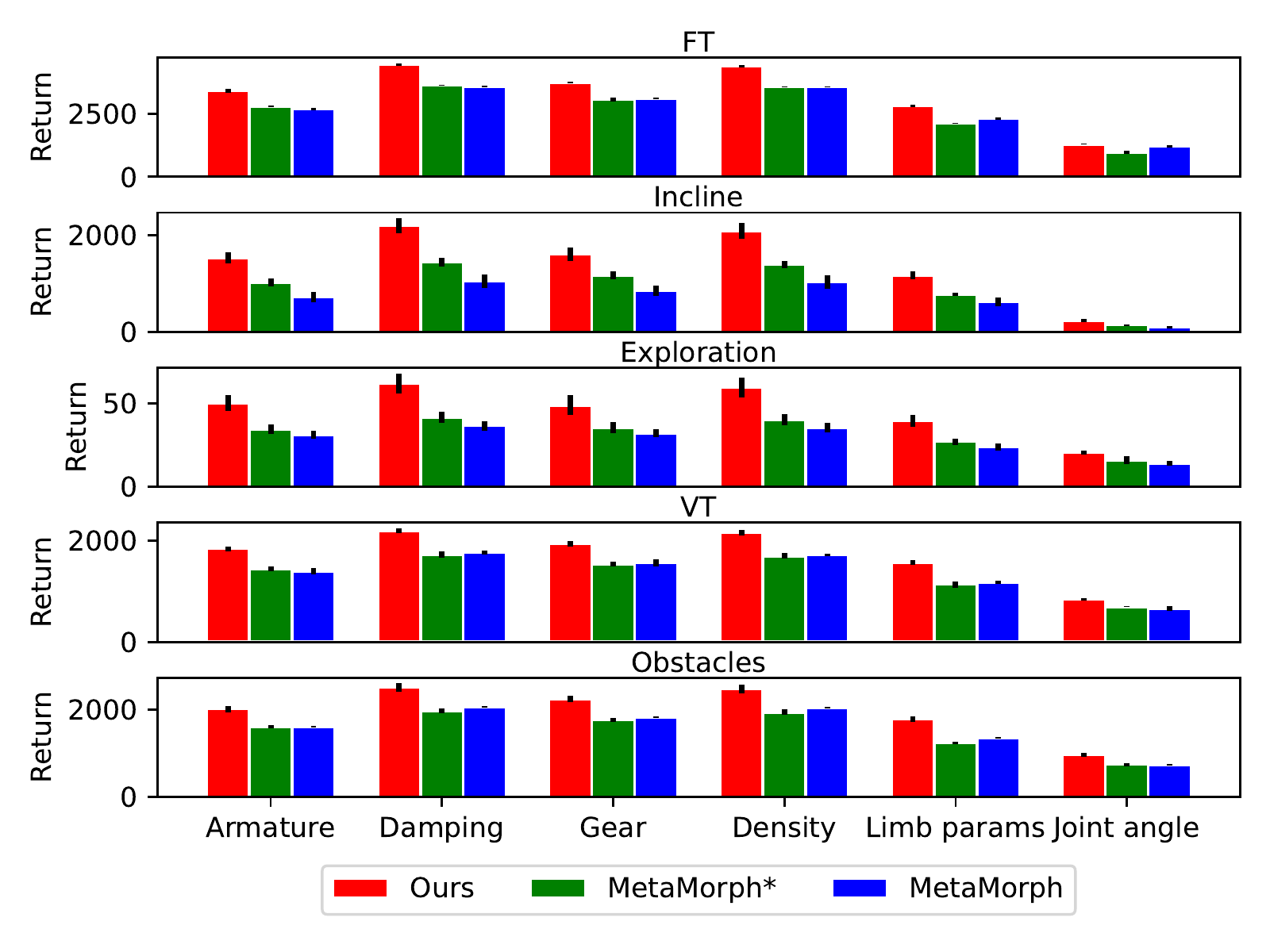}
    \caption{Zero-shot generalization performance of different methods to kinematics and dynamics variations. The rows correspond to the 5 environments, and the columns correspond to parametric variations in 6 different morphology context parameters. }
    \label{fig:eval_variation}
\end{figure}
As shown in Figure \ref{fig:eval_variation}, our method consistently outperforms the baselines, with an average improvement ratio of 26\%, 50\%, 43\%, 28\%, 30\% in each environment compared to MetaMorph*, which validates that our method not only enables better multi-robot training, but also generalizes better to unseen robots with parametric variations.
However, we also notice that zero-shot generalization to kinematics variation (especially joint angles) is much harder, as is also reported in \citet{gupta2022metamorph}. This is mainly because that changes in joint angles may significantly influence the feasible actions and the gait for locomotion, and how to tackle this challenge is an interesting direction for future work.

\subsection{Zero-Shot Generalization to Unseen Morphologies}

Table \ref{table:generalization} shows the zero-shot generalization performance of different methods in each environment, which generally follows the same trend as during training, i.e., the model with higher training scores also performs better in zero-shot generalization. 
Specifically, our method outperforms MetaMorph* by $18\%$, $29\%$, $24\%$, $27\%$ and $37\%$ in each environment respectively. 
This implies that our contextual modulation method can indeed better model the dependence of the control policy on the robot morphology, instead of simply overfitting to the training morphologies via its more complicated architecture designs. 
However, the large variance in the return across different seeds does imply that improving zero-shot generalization on unseen morphologies is still an open problem for future work.
\begin{table*}
\caption{Zero-shot generalization performance of different methods to test morphologies with unseen topology graphs. The best method in each environment is marked in \textbf{Bold}. The methods that are not statistically significantly different from the best method are marked by \underline{underline} based on Welch’s t-test with a significance level of 0.05 \citep{colas2019hitchhiker}. } 
\label{table:generalization}
\vskip 0.15in
\begin{center}
\begin{small}
\begin{sc}
\begin{tabular}{lccccc}
\toprule
Environment & MetaMorph & MetaMorph* & FA & HN & FA+HN \\
\midrule
FT & $\underline{1384\pm62}$ & $1266\pm105$ & $\underline{1439\pm27}$ & $1259\pm112$ & $\bf{1490\pm59}$ \\
Incline & $27\pm32$ & $\underline{312\pm136}$ & $\bf{468\pm58}$ & $\underline{312\pm97}$ & $\underline{403\pm66}$ \\
Exploration & $\underline{19\pm1}$ & $\underline{19\pm1}$ & $\underline{22\pm2}$ & $16\pm3$ & $\bf{23\pm3}$ \\
VT & $\underline{752\pm62}$ & $\underline{767\pm23}$ & $\underline{860\pm112}$ & $\underline{900\pm24}$ & $\bf{971\pm122}$ \\
Obstacles & $866\pm30$ & $829\pm50$ & $937\pm46$ & $969\pm47$ & $\bf{1133\pm12}$ \\
\bottomrule
\end{tabular}
\end{sc}
\end{small}
\end{center}
\vskip -0.1in
\end{table*}

\subsection{Qualitative Analysis}

We conduct a qualitative analysis in the FT environment to illustrate the difference in the locomotion skills learned by different methods. 
\begin{figure}
    \centering
    \includegraphics[width=0.8\columnwidth]{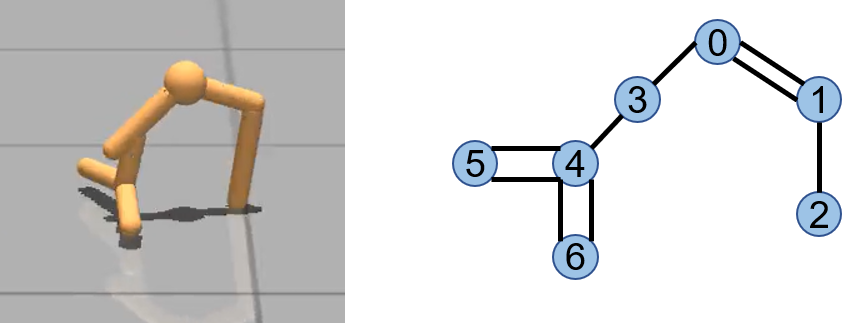}
    \caption{The example morphology and its morphology tree. Some limbs are connected to their parents via two joints, represented by the two edges between nodes. The sphere node is the torso of the robot and also the root of the morphology tree. }
    \label{fig:example_morphology}
\end{figure}

\begin{figure}
    \centering
    \includegraphics[width=\columnwidth]{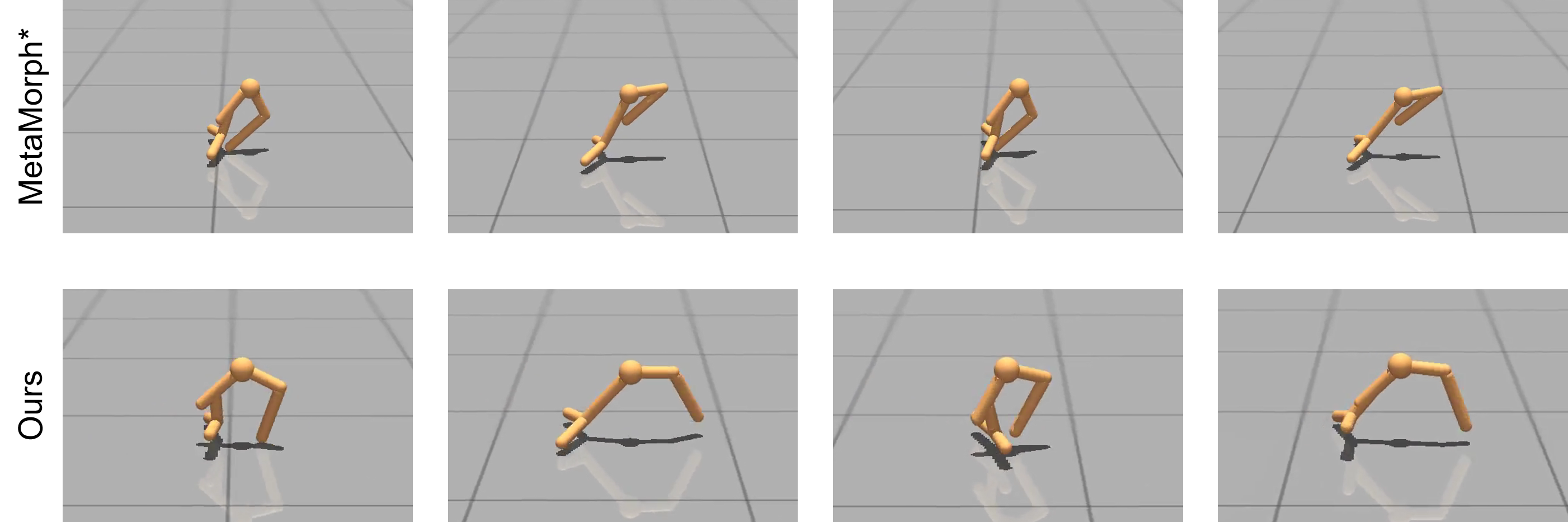}
    \caption{Visualization of the locomotion trajectories learned by MetaMorph* and our method on the example robot.}
    \label{fig:locomotion}
\end{figure}

We experiment on an example morphology as shown in Figure \ref{fig:example_morphology}, and compare the locomotion learned by MetaMorph* and our method in Figure \ref{fig:locomotion}. 
For MetaMorph*, the robot moves forward by kicking the ground with its front limb. 
However, the front limb does not fully stretch out, thus provides limited forward force and makes the body unstable. 
In 1000 timesteps, the robot falls twice and only achieves a return of 1375 in its best trial. 
By contrast, our method learns a policy that fully stretches out the front limb to provide stronger forward force, and better coordinates the movement of the front and back limbs. 
The robot runs more stably without any failure during evaluation, and achieves a much higher return of 4612. 

\begin{figure}
    \centering
    \includegraphics[width=\columnwidth]{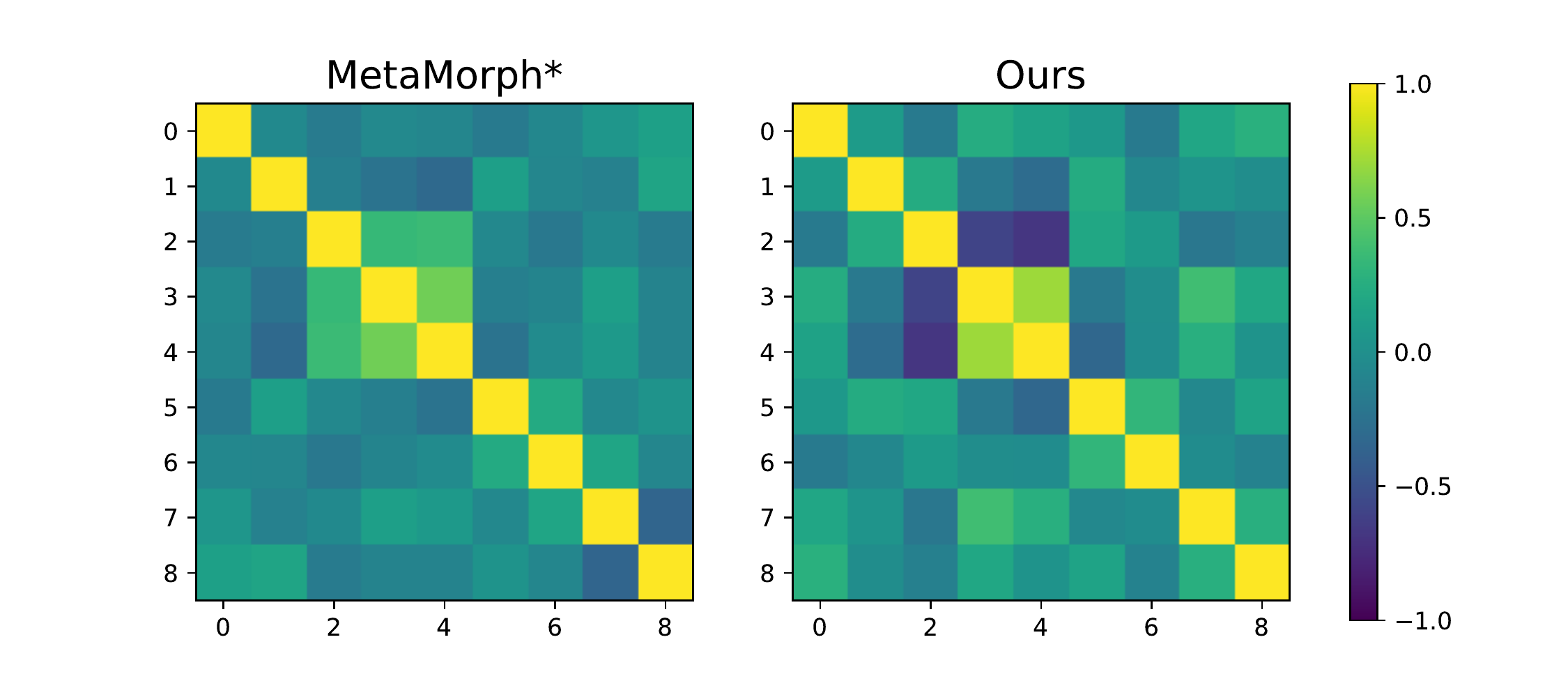}
    \caption{The correlation matrix of different action dimensions on the example morphology.}
    \label{fig:action_correlation}
\end{figure}

In addition to behavior visualization, we further analyze the correlation between the action sequences taken by different limbs as an indicator of behavior synergies. 
Intuitively, if the behavior of two limbs are better coordinated, then we may expect their action sequences to have a higher correlation coefficient. 
Figure \ref{fig:action_correlation} shows the correlation matrix between different action dimensions on the example morphology. 
For our method, joint 2 (which controls the front limb 2) is much better synchronized with joints 3 and 4 (which control limbs 3 and 4 in the back). 
This reflects how the periodic gait of our method in Figure \ref{fig:locomotion} is generated. 

In the VT environment, MetaMorph* has an average action correlation of 0.24 across all training morphologies, while our method has 0.29. 
This higher correlation indicates a better synergy between limbs, which may help explain why our method has more fluent and stable locomotion. 

\section{Related Work}
\paragraph{Universal Morphology Control}
To learn a universal policy to control multiple robots, many previous works focus on the setting where the robots share the same morphology and only differ in kinematics or dynamics parameters \citep{chen2018hardware,peng2018sim,clavera2018learning,ghadirzadeh2021bayesian,feng2022genloco}. 
These works mainly build upon MLP architectures, thus cannot handle different morphologies with heterogeneous state and action spaces. 
\citet{wang2018nervenet}, \citet{pathak2019learning} and \citet{huang2020one} use GNNs \citep{wu2020comprehensive} to tackle this problem, as the robot morphology can be seen as a kinematic graph and GNNs can naturally generalize across graphs with different number of nodes.
\citet{kurin2021my} show that it is hard to model the interactions between distant nodes in the morphology graph with GNNs, thus propose to use transformers as the controller to enable immediate interactions between any node pairs. 
While these works mainly focus on architecture design to better model limb interactions, more recent works show that incorporating morphology information into the controller via feature concatenation or positional encoding can further improve learning performance \citep{gupta2022metamorph,hong2022structureaware,trabucco2022anymorph}. 
However, these approaches in effect just add a context-conditioned bias term to the node embedding, which may not be sufficient to model the complex dependency of a robot's control policy on its morphology. 
Moreover, instead of sharing all parameters across different morphologies, learning node-wise or morphology-wise parameters for some specific layers has also been shown to improve learning performance \citep{d2020sharing,yuan2022transformact}, but suffers from generalization and scalability issues as discussed in Section \ref{subsec:HN}. 
Unlike existing works, the contextual modulation method in this paper enables both learning diverse morphology-conditioned policies, and generalization and scalability to new robots. 

\paragraph{Contextual Modulation in RL}
The optimal policy for a task usually critically depends on the task context that defines the task's characteristics.
Consequently, conditioning the policy on the task context may significantly improve its training performance and generalization ability over a distribution of tasks compared to context-agnostic learning \citep{benjamins2022contextualize}. 
To learn a context-conditioned policy, an importance design choice is the architecture used to incorporate the task context into the policy, which reflects our inductive bias on the task structure. 
Instead of simply concatenating the context features to the state features, which is limited in model capacity \citep{galanti2020modularity}, 
different architectures have been proposed to modulate the policy via task context, such as feature-wise multiplication \citep{ben2022context,benjamins2022contextualize}, a routing network that determines how to combine different skill modules for a specific task \citep{yang2020multi,sodhani2021multi,ponti2022combining}, and hypernetworks \citep{yu2019multi,peng2021linear,sarafian2021recomposing,beck2022hypernetworks,rezaei2022hypernetworks}.
Our work shares a similar motivation as these methods, but focuses on a more challenging domain of universal morphology control where different tasks do not share the same state and action space. 

\section{Conclusion}
In this paper, we investigated how to learn a universal control policy for different robot morphologies. 
To better model the dependency of the control policy on the robot morphology, we proposed a hierarchical architecture to modulate the base controller with morphology context, which includes a hypernetwork module that generates morphology-dependent controller parameters, and a morphology-dependent attention module to modulate the transformer layers in the base controller. 
Experimental results validated the effectiveness of our method on both multiple training robots and unseen test morphologies. 

For future work, an interesting direction is how to learn better context representation for modulation. 
In this paper, we directly used the original node context features provided in the benchmark as the modulator input, and a simple MLP as the context encoder. 
How to design better context features, such as utilizing node connectivity information, and how to design better context encoding architectures are both interesting topics to investigate. 
Another potential direction is how to improve zero-shot generalization performance on unseen robots, as there is still a large gap between the current generalization results and the optimal performance we can achieve by directly training on the test robots. 
Thirdly, our method builds upon a modular design space assumption which may not hold on some real-world robots, thus how to relax this assumption to enable more general knowledge transfer across different morphologies is an interesting direction for future work. 
Finally, while we focus on the problem setting of learning a universal controller over a set of pre-given robot morphologies, an interesting direction for future work is to apply our method to a closely related problem setting of jointly optimizing the morphology design and its corresponding control policy \citep{schaff2019jointly,wang2018neural,yuan2022transformact,schaff2022n}. 

\section*{Acknowledgements}
We would like to thank Zhengdao Chen, Agrim Gupta, Matthew Jackson, Vitaly Kurin and Risto Vuorio for their helpful discussion on the work. 
We would also like to thank the conference reviewers for their constructive feedback on the paper. 
Zheng Xiong is supported by UK EPSRC CDT in Autonomous Intelligent Machines and Systems (grant number EP/S024050/1) and AWS.
Jacob Beck is supported by the Oxford-Google DeepMind Doctoral Scholarship. The experiments were made possible by a generous equipment grant from NVIDIA.

\bibliography{reference}
\bibliographystyle{icml2023}

\newpage
\appendix
\onecolumn
\section{Implementation Details}
\label{appendix:implementations}

\subsection{Architecture Details of Contextual Modulation}
Intuitively, using GNNs or transformers as the context encoder to model node interactions may learn better context representations. 
However, in practice we find that using simple MLPs as the context encoder achieves similar or even better performance. 
The reason might be that we have many fewer training samples for the context encoder, which equals the number of robots we have for training, as the morphology context does not change on a robot.  
So using models with high capacity may not be helpful here and even lead to overfitting. 
Moreover, HNs are known to be hard to optimize \citep{chang2019principled}, and we find that using transformers as the context encoder makes HN training unstable. 
Consequently, we just use MLPs as the context encoder shared over different nodes. 
Specifically, we train two separate context encoders for HN and FA respectively. The context encoder is a 2-layer MLP for HN, and a 3-layer MLP for FA, both with 128 units in each hidden layer. 

The HN output layer is implemented as a linear mapping from context encoding to the parameters in the base network, with one independent output head for each modulated layer in the base controller. 
We initialize it with the \emph{Bias-HyperInit} method proposed by \citet{beck2022hypernetworks}, i.e., the weights are set to 0, and the biases are sampled from the same distribution that is used to initialize the modulated layer in the base network. 
In this way, all the nodes share the same control parameters just like MetaMorph at the beginning, and gradually develop node-wise diversity while the HN weights are updated. 

\subsection{Proprioceptive and Context Features}

We use the same proprioceptive and context features as in MetaMorph for a fair comparison, which can be found in Appendix A.1 of \citet{gupta2022metamorph}.

\section{PPO and Early Stopping}
\label{appenfix:early_stop}

PPO \citep{schulman2017proximal} is an on-policy RL algorithm that takes multiple steps of update on the current data we have, while  also trying not to exceed some trust region boundary to avoid performance collapse. 
For a state-action pair $s, a$, the clipping objective function of PPO is defined as $$ L(s,a,\theta_k, \theta) = \min \left(\frac{\pi_\theta(a|s)}{\pi_{\theta_k}(a|s)} A^{\pi_{\theta_k}}(s,a), \text{clip}\left( \frac{\pi_\theta(a|s)}{\pi_{\theta_k}(a|s)}, 1-\epsilon, 1+\epsilon \right) A^{\pi_{\theta_k}}(s,a) \right), $$
where $\pi_{\theta_k}$ is the behavior policy used to collect on-policy data for policy update in iteration $k$, $\pi_{\theta}$ is the target policy we want to learn, $A^{\pi_{\theta_k}}(s,a)$ is the advantage function of $\pi_{\theta_k}$, and $\epsilon$ is a hyperparameter that constrains the updated policy to not be too far away from the behavior policy. 

In many RL libraries, one PPO iteration is implemented by first collecting $N$ steps of rollout data on $M$ workers in parallel with $\pi_{\theta_k}$, then dividing the collected data into $B$ batches and repeating minibatch update for $T$ epochs. 
Consequently, for one PPO iteration, we'll do $T\cdot B$ times of minibatch update. 

However, the clipping objective function alone can not guarantee policy update within the trust region. 
Early stopping is a solution to this problem by first checking the approximate KL divergence between $\pi_\theta$ and $\pi_{\theta_k}$ before each minibatch update, and terminating the current update iteration if the divergence exceeds a threshold value $\delta$. 
The policies' KL divergence is approximated as $D_{\text{KL}}(\pi_{\theta_k} || \pi_\theta) \approx \sum_{(s,a)\in\mathcal{B}} \log \left( \frac{\pi_{\theta_k}(a|s)}{\pi_\theta(a|s)} \right) $, where $\mathcal{B}$ is the current data minibatch used for policy update. 

The early stopping threshold $\delta$ is a critical hyperparameter in PPO, as it gives a measurement of the range of the trust region we allow for policy update. 
We tune it over the candidate set of $\{0.03, 0.05\}$, and report the optimal value of $\delta$ for each method in each environment in Table \ref{table:ES}. 
We do not tune over a wider range, as empirically we found that a even smaller value of $\delta$ usually causes early stopping to happen too early, while a larger value allows for a too large trust region, both harming the learning performance. 

\begin{table*}[t]
\caption{Optimal value of the early stopping threshold for each method in each environment. }
\label{table:ES}
\vskip 0.15in
\begin{center}
\begin{small}
\begin{sc}
\begin{tabular}{lcccc}
\toprule
Environment & MetaMorph* & FA & HN & FA+HN \\
\midrule
FT & 0.05 & 0.05 & 0.05 & 0.05 \\
Incline & 0.03 & 0.05 & 0.05 & 0.05 \\
Exploration & 0.03 & 0.03 & 0.03 & 0.03 \\
VT & 0.03 & 0.03 & 0.03 & 0.03 \\
Obstacles & 0.03 & 0.03 & 0.03 & 0.03 \\
\bottomrule
\end{tabular}
\end{sc}
\end{small}
\end{center}
\vskip -0.1in
\end{table*}

\section{Analysis on MetaMorph}
\label{appenfix:metamorph}

In this section, we introduce the issues found when reproducing MetaMorph results with its source code, which motivates us to propose the MetaMorph* variant as an alternative baseline. 

The MetaMorph paper reports significant improvement in learning performance by adding positional encoding (PE) to the node embedding. 
PE is a common practice in transformers to incorporate positional information into the embedding of each element \citep{vaswani2017attention}. 
Specifically, MetaMorph adopts learned PE, i.e., the PE for each position is a vector that is learned during training instead of hard-coded in advance. 
However, a robot morphology is structured as a tree, which does not hold sequential information about each node by nature. 
So MetaMorph first traverses each morphology tree via depth-first search to turn it into a 1D sequence, then index each node by its position in the sequence. 
One PE vector is learned for each position in the sequence, and the nodes with the same index across different robots will share the same PE vector. 

\begin{figure}
    \centering
    \includegraphics[width=0.8\textwidth]{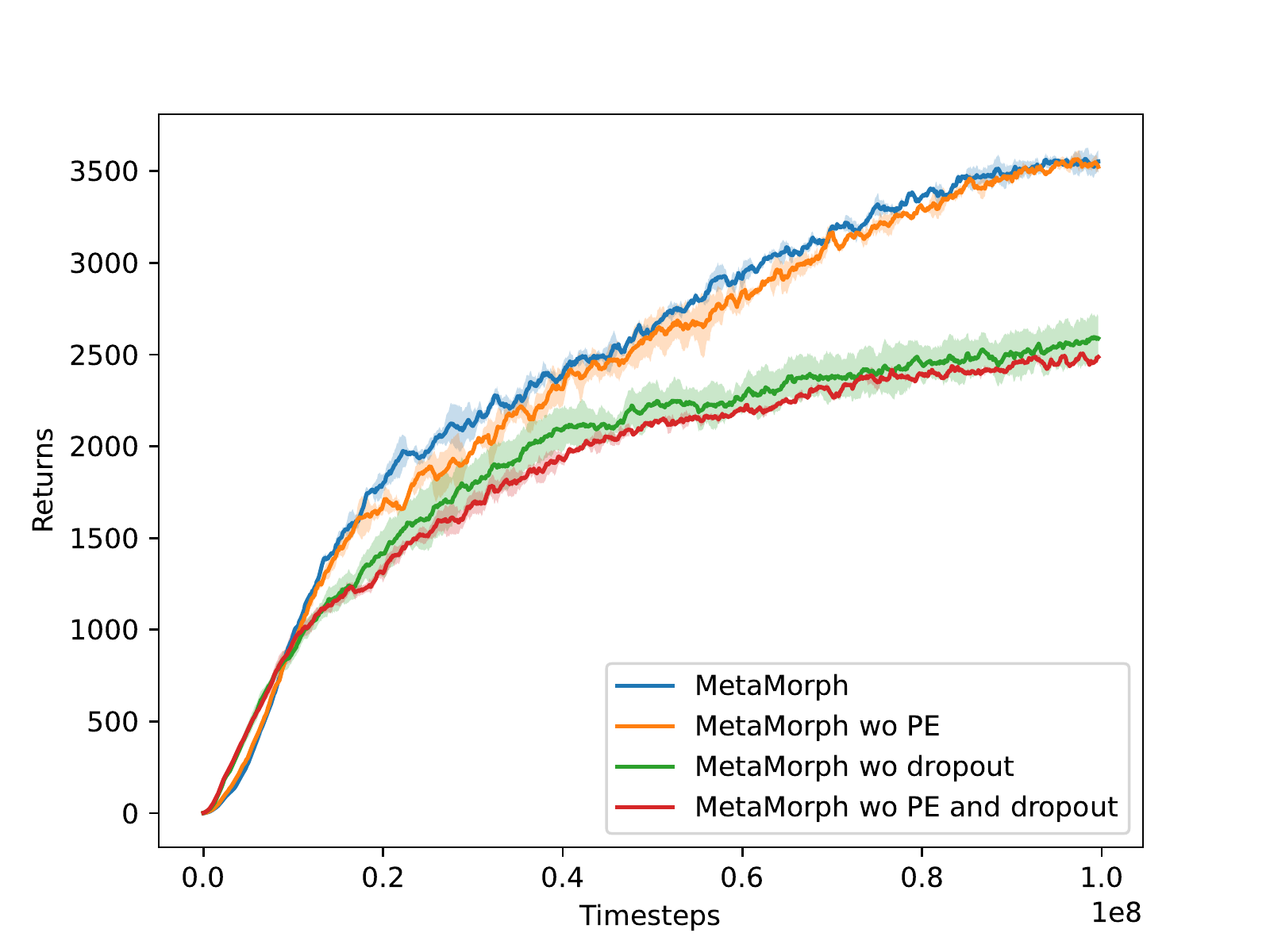}
    \caption{The effect of PE and dropout on the performance of MetaMorph in the FT environment. }
    \label{fig:PE_analysis}
\end{figure}

However, we found that in the source code of MetaMorph, PE is implemented as $ e_i'=\text{dropout}(e_i+\text{PE}_i)$, where $e_i$ is the embedding of node $i$. 
To investigate which operation actually contributes to the performance improvement, we experiment with $e_i'=\text{dropout}(e_i)$ and $e_i'=e_i+\text{PE}_i$ respectively, and surprisingly find that the dropout operation is the main contributor here, while PE alone makes little difference in training performance (Figure \ref{fig:PE_analysis}). 

\subsection{Why PE Does Not Help?}
\label{appendix:PE_analysis}
\begin{figure}[t]
    \centering
    \includegraphics[width=0.8\textwidth]{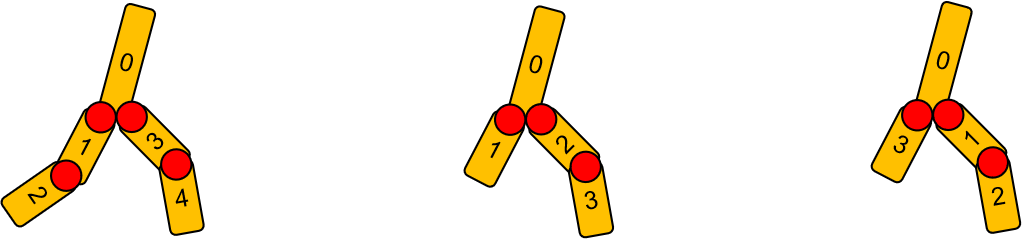}
    \caption{Sources of inconsistency in PE across morphologies. (1) The nodes that play different roles in different morphologies may share the same PE, such as the two nodes indexed by 2 in the left and middle robot. (2) There is no intrinsic order between the children of a parent node in the robot morphology, so PE is sensitive to how we choose which child node to expand first, such as the middle and right robot which have the same morphology but totally different PE for each non-root node. }
    \label{fig:PE_issues}
\end{figure}

Our hypothesis here is that PE has the benefit of enabling more diverse behaviors across different nodes, but also has the drawback of adding the same PE vector to nodes with different physical meanings across robots, i.e., PE is not consistent across morphologies (Figure \ref{fig:PE_issues} shows two sources of inconsistency in PE). 
And when training on multiple robots, the  drawback outweighs the benefit, so PE provides no performance gain overall. 

\begin{figure}
    \centering
    \includegraphics[width=0.8\textwidth]{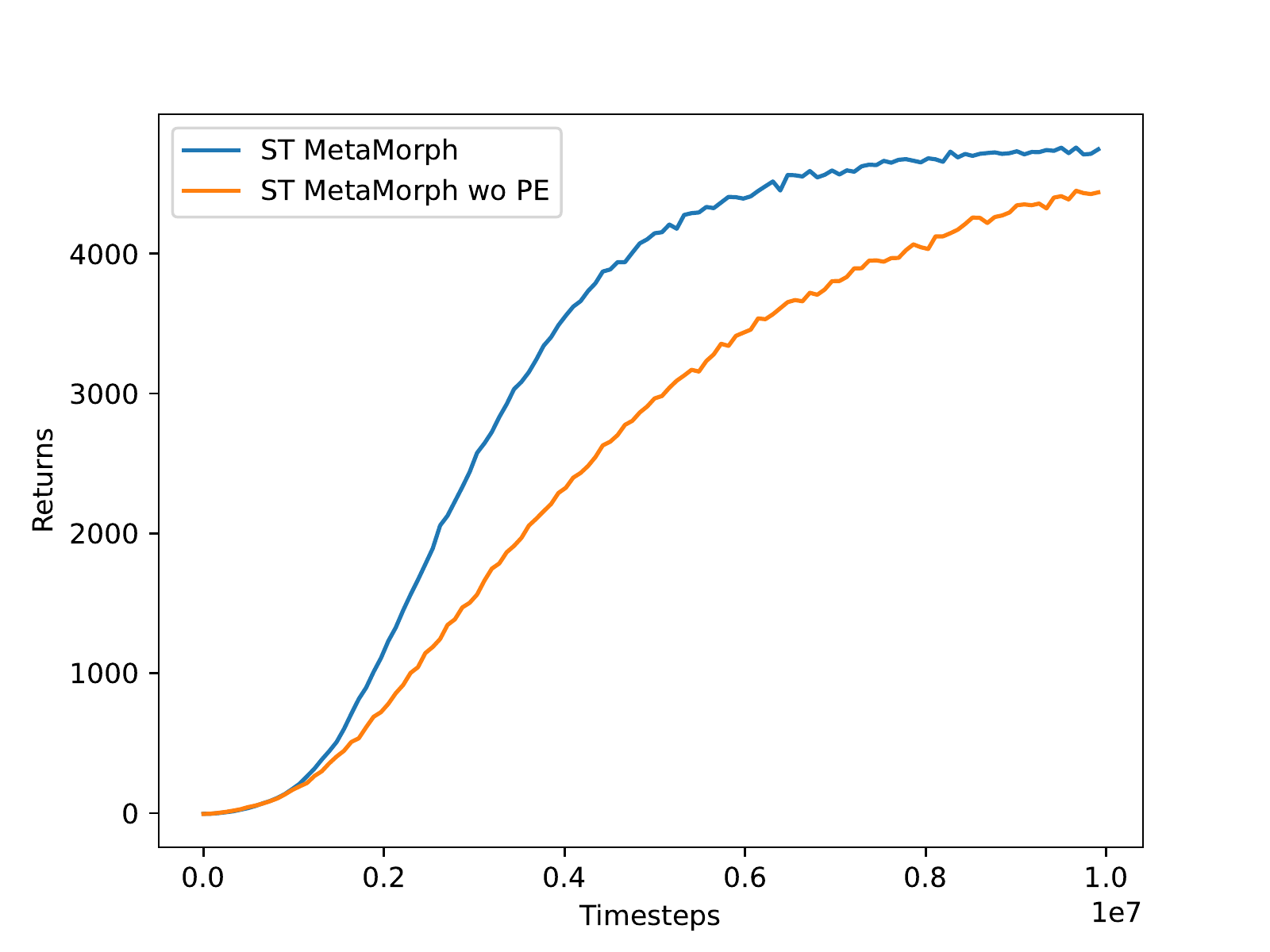}
    \caption{The effect of PE on single-task MetaMorph in the FT envorionment. }
    \label{fig:ST_PE}
\end{figure}

To validate this hypothesis, we investigate the effect of PE in single-task (ST) training. 
If our hypothesis holds, we should observe better performance by using PE compared to not, as there is no inconsistency issue on a single robot, while the benefits of PE maintains. 
We experiment on 30 morphologies and show their avergage learning curve in Figure \ref{fig:ST_PE}. 
As expected, PE indeed improves training performance in ST training, which proves that using PE to distinguish between different nodes in a single morphology is helpful. 
However, the inconsistency issue breaks its effectiveness in the multi-morphology training setting. 

Based on the above analysis, we conclude that the PE implementation in MetaMorph is not essential for good performance, thus decide to not include it in MetaMorph*. 

\subsection{Why Dropout Helps?}

It's quite surprising that removing the dropout operation causes such a significant performance drop in MetaMorph, as dropout is not believed to be a very useful regularizer for on-policy RL algorithms \citep{liu2021regularization}. 
Furthermore, the dropout operation is implemented in an inconsistent way in MetaMorph, which introduces significant noise to the action probability ratio $r=\frac{\pi_\theta(a|s)}{\pi_{\theta_k}(a|s)}$. 
Specifically, if a dropout mask $m$ is applied to a state $s$ during data collection, then the same mask should be used when $s$ is used during policy update, i.e., $r=\frac{\pi_\theta(a|s; m)}{\pi_{\theta_k}(a|s; m)}$, to maintain a consistent ratio computation. 
However, in the MetaMorph code, a different dropout mask $m'$ is randomly sampled whenever $s$ is used for policy update, i.e., $r'=\frac{\pi_\theta(a|s; m')}{\pi_{\theta_k}(a|s; m)}$, which introduces significant noise. 
For example, before the first minibatch update in a PPO iteration, we expect $r$ to be 1 for each state-action pair in the minibatch, as $\pi_\theta = \pi_{\theta_k}$ before policy update. 
However, using inconsistent dropout masks will make $r$ unequal to 1 even before any policy update, which does not make sense intuitively. 

\begin{figure}
    \centering
    \includegraphics[width=0.8\textwidth]{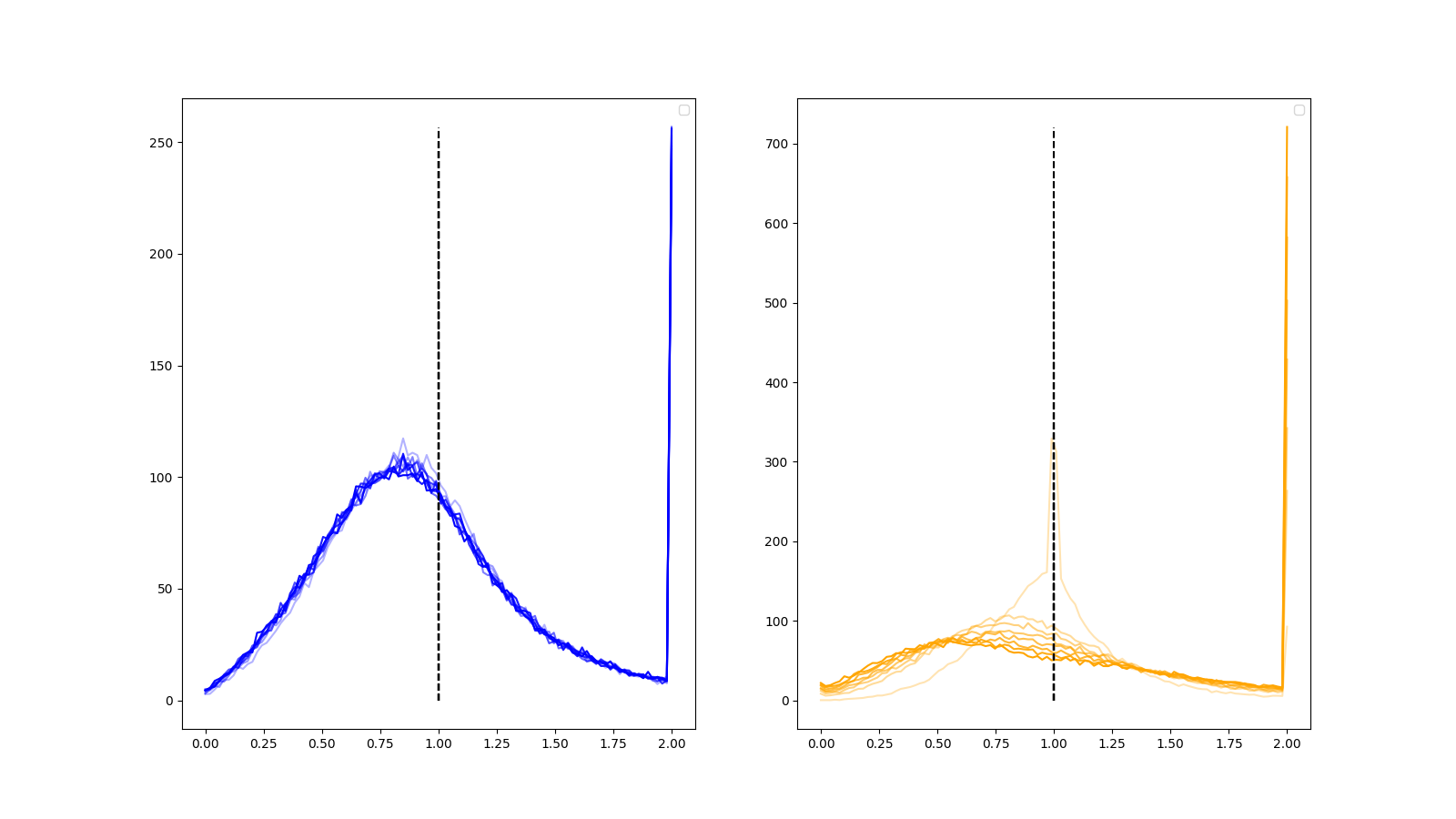}
    \caption{Ratio distribution of each epoch during one PPO update iteration. Left: MetaMorph; Right: MetaMorph without dropout. Lighter color represents earlier epoch during one update iteration. There is a spike in the right of each subplot because we clip the ratios to be within $[0,2]$.}
    \label{fig:dropout}
\end{figure}

We thus look deeper into the training statistics of PPO to better understand why this inconsistent dropout operation works, and notice that using dropout or not causes a significant difference in the distribution shift of $r$ during the learning process. 
Figure \ref{fig:dropout} shows how the distribution of $r$ changes on each epoch during one PPO update iteration. 
We can see that the inconsistent dropout operation somehow maintains the distribution stable across epochs, while the distribution significantly changes if learning without dropout. 
This implies that when learning without dropout, the policy has very likely exceeded the trust region and thus performs worse. 

A natural solution to restricting the distribution shift is the early stopping method proposed in Appendix \ref{appenfix:early_stop}. 
MetaMorph actually already uses ES in its code, but the threshold is set to 0.2, which is too large and in practice seldomly triggers early stopping. 
We set it to a smaller value of 0.03 or 0.05, so that we can achieve similar performance as MetaMorph without using the inconsistent dropout operation, which is the second modification we make in MetaMorph*. 


\end{document}